%% file: samplepaper.tex
\documentclass[runningheads]{llncs}
\usepackage{graphicx}
\usepackage{graphicx}
\usepackage{comment}
\usepackage{amsmath}
\usepackage{amssymb}
\usepackage{tikz-cd}
\usepackage[makeroom]{cancel}
\usepackage{longtable}
\usepackage[colorlinks=true,linkcolor=blue,citecolor=green]{hyperref}
\usepackage[margin=3.5cm]{geometry}
\usepackage{stmaryrd}
\usepackage{bussproofs}
\usepackage{mathtools}
\usepackage{mathrsfs}
\usepackage{multirow}
\usepackage[all]{xy}
\usepackage[colorinlistoftodos,prependcaption,textsize=small]{todonotes}
\input{macros}

\begin{document}
\title{Updating belief functions over Belnap--Dunn logic\thanks{The research of Sabine Frittella and Sajad Nazari was funded by the grant ANR JCJC 2019, project PRELAP (ANR-19-CE48-0006).}}

\author{
Sabine Frittella\inst{1}\orcidID{0000-0003-4736-8614} \and
Ondrej Majer\inst{2}\orcidID{0000-0002-7243-1622}
\and
Sajad Nazari\inst{3}\orcidID{0000-0002-4295-2435} }
\authorrunning{S. Frittella, O. Majer and S. Nazari}

\institute{INSA Centre Val de Loire, Univ.\ Orl\'{e}ans, LIFO EA 4022, France\\
\email{sabine.frittella@insa-cvl.fr,sajad.nazari@insa-cvl.fr}
\and
Czech Academy of Sciences, Institute of Philosophy, 
Prague\\
\email{majer@flu.cas.cz}}
\maketitle              
\begin{abstract}
Belief and plausibility are weaker measures of uncertainty than that of pro\-ba\-bility. They are motivated by the situations when full probabilistic information is not available. However, information can also be contradictory. Therefore, the framework of classical logic is not necessarily the most adequate. Belnap—Dunn logic was introduced to reason about incomplete and contradictory information. \cite{BilkovaFrittellaKozhMajerNazari} and \cite{KleinMajerRad2021} generalize the notion of probability measures and belief functions to Belnap—Dunn logic, respectively. In this article, we study how to update belief functions with new pieces of information. We present a first approach via a frame semantics of Belnap—Dunn logic. 
\end{abstract}
\section{Introduction}

Belief functions were introduced to generalise the notion of probabilities to situations with incomplete information. They can be used to encode the information given by one or many pieces of evidence. They were first introduced on Boolean algebras \cite{shafer1976}, that is within the framework of classical reasoning. 
Dempster--Shafer theory uses the equivalent representation of belief functions via their mass functions to propose a method to aggregate the information conveyed by two belief functions.
Since, the theory of belief functions has been developed  on distributive lattices \cite{barthelemy,Grabisch2008,Zhou13} and finite lattices \cite{FMPTW20}. 
Notice that the definition of belief functions (see Definition \ref{def:generalbelieffunction}) relies only on the lattice structure of the algebra. However, its dual notion, plausibility, is usually defined by combining a belief function and the classical negation. In this framework, belief is interpreted as a lower bound on the probability of an event and plausibility as an upper bound. Similarly, Dempster--Shafer combination rule also relies strongly on the underlying algebra being Boolean.
In this article, we discuss the interpretation of belief and plausibility functions within the framework of Belnap--Dunn logic rather than through classical logic, because it allows us to reason with incomplete and contradictory information.

Belnap--Dunn logic was introduced to reason about information rather than about truth. In classical logic a statement $p$ is either \textit{true} or \textit{false}, meaning that $p$ is \textit{true} (resp.\ false) iff the statement $p$ is true (resp.\ false) in the world. 
In Belnap--Dunn ($\BD$) logic, a statement $p$ is either ``supported by the available information", or ``contradicted by the available information", or ``neither supported nor contradicted by the available information", or ``both supported and contradicted by the available information". These four truth values are respectively denoted $\mathbf{t}$ (\textit{true}), $\mathbf{f}$ (\textit{false}), $\mathbf{n}$ (\textit{neither}), $\mathbf{b}$ (\textit{both}). 
\cite{BilkovaFrittellaKozhMajerNazari} introduces belief functions over $\BD$ models and presents a logic to reason with them. This work was motivated by the counterintuitive results that Dempster--Shafer ($\DS$) theory can produce in presence of highly contradictory pieces of information.
Indeed, the strategy of Dempster--Shafer combination rule is to ignore  contradictory information rather than dealing with the contradictions and saying something meaning full about them. Since $\BD$ logic is a simple, well-established extension of classical logic that was introduced to formalised reasoning based on incomplete and contradictory information, it appeared natural to adapt Dempster--Shafer theory over $\BD$ logic.

The next step is to look at taking new pieces of information into account by updating the belief functions.
\cite{Halpern2017} presents different ways to update belief and plausibility functions in the classical framework.
This article expends on \cite{BilkovaFrittellaKozhMajerNazari} and discusses how to adapt and interpret those results within the framework of belief and plausibility functions over Belnap--Dunn logic. 

\paragraph{Structure of the paper.} In Section \ref{sec:prelim}, we present $\BD$ logic and non-standard probabilities over $\BD$ logic, then we recall  definitions and lemmas about belief and plausibility functions and Dempster--Shafer combination rule, finally we present existing proposals to update belief functions over Boolean algebras. In Section \ref{sec:update}, we introduce models for belief and plausibility over $\BD$ logic.
We discuss how to update belief and plausibility when getting a new piece of information.
In Section \ref{sec:conclusion}, we discuss further research.

\section{Preliminaries}
\label{sec:prelim}

In this section, we first introduced the necessary definitions about $\BD$ logic and non-standard probabilities. Then we recall useful definitions and lemmas about belief and plausibility  functions and Dempster--Shafer combination rule. Finally, we present two ways to update belief functions on Boolean algebras.

In the reminder of the paper, we will always work with finite lattices. Recall that a \emph{lattice} is a tuple $\mainL=\langle L,\vee,\wedge\rangle$, such that 
$\vee$ and $\wedge$ are binary, commutative, associative, and idempotent operations that satisfy the following rules: $x\vee (x\wedge y) = x$ and $x\wedge (x\vee y) = x$, for all $x,y\in\mainL$.
A lattice is \emph{bounded} if it contains nullary operators $\bot$ and $\top$ that represent respectively its lower and upper bounds, i.e., 
for every element $x\in\mainL$, we have $x\vee\top=\top$ and $x\wedge\bot=\bot$. 
A lattice is \emph{distributive} if $(x\vee y)\wedge z=(x\wedge z)\vee(y\wedge z)$ holds for all $x,y,z\in\mainL$.
A \emph{(bounded) De Morgan algebra} is a (bounded) distributive lattice equipped with an additional unary operation $\neg$ such that $\neg\neg x=x$ and $\neg(x\wedge y)=\neg x\vee\neg y$ for all $x,y\in\mathcal{L}$.
A \emph{Boolean algebra} is a bounded De Morgan algebra that satisfies the law of excluded middle ($p \vee \neg p =\top$) and the principle of explosion ($p \wedge \neg p =\bot$). A function  $\mu:\PS\rightarrow[0,1]$ is called a \emph{(finitely additive) probability measure} if it satisfies the following properties:
(i) $\mu(S)=1$, and
(ii) $\mu(A \cup B) = \mu(A) + \mu (B)$ for $A, B$ disjoint.

\subsection{Belnap--Dunn logic}
 
$\BD$ logic mentioned in the introduction was introduced by Nuel Belnap in~\cite{Belnap2019}. His main aim was to design a logical  system capable of dealing with inconsistent or/and incomplete information.  
The language $\LBD$ of $\BD$ logic is defined via the following grammar over a finite set of propositions $\Prop$:  
\[\ \phi\coloneqq p\in\Prop\mid\neg\phi\mid \phi\wedge\phi \mid \phi\vee\phi \mid \bot\mid \top.\]

The constants $\bot$ and $\top$ are not a standard part of the signature of $\LBD$,  
we include them for the sake of simplicity. Most of the  results can be straightforwardly adapted for the more general framework without the constants. 
Semantics for $\BD$ logic can be provided in two equivalent ways. One possibility is to evaluate the formulas directly into  the set $\{\mathbf{t}, \mathbf{f}, \mathbf{n}, \mathbf{b}\}$. We will use the second option based on the idea of  independence of positive and negative information. In particular, a lack of positive support for a claim does not automatically mean a support for its negation. This is formally represented by two valuations representing positive (negative) support, respectively.

\begin{definition}[Belnap--Dunn models]
A \emph{Belnap--Dunn model  ($\BD$ model)} is a tuple $\mathfrak{M}=\langle S,v^+,v^-\rangle$ where $S\neq\varnothing$ is a finite set of states and $v^+,v^-:\Prop\rightarrow {\mathcal{P}}(S)$ are positive, negative valuation functions respectively.
\end{definition}
The valuations are extended to the corresponding satisfaction relations. Each of them evaluates compound formulas in a way which is in fact classical: e.g. a conjunction is supported positively in a state iff each of its conjuncts is, and it is supported negatively if at least one of the conjuncts is.
\begin{definition}[Frame semantics for $\BD$]\label{def:BDframesemantics}
Let $\phi,\phi'\in\LBD$,  $\mathfrak{M}=\langle W,v^+,v^-\rangle$ a $\BD$ model and $w\in S$. Then $\vDash^+$ and $\vDash^-$ are defined as follows.
\begin{align*}
w\vDash^+p&\text{ iff }w\in v^+(p)&w\vDash^-p&\text{ iff }w\in v^-(p)\\
w\vDash^+\neg\phi&\text{ iff }w\vDash^-\phi&w\vDash^-\neg\phi&\text{ iff }w\vDash^+\phi\\
w\vDash^+\phi\wedge\phi'&\text{ iff }w\vDash^+\phi\text{ and }w\vDash^+\phi'&w\vDash^-\phi\wedge\phi'&\text{ iff }w\vDash^-\phi\text{ or }w\vDash^-\phi'\\
w\vDash^+\phi\vee\phi'&\text{ iff }w\vDash^+\phi\text{ or }w\vDash^+\phi'&w\vDash^-\phi\vee\phi'&\text{ iff }w\vDash^-\phi\text{ and }w\vDash^-\phi'
\end{align*}
\end{definition}

We will make use of the notions of the  \emph{positive extension of a formula} (the set of states supporting it): $|\f|^+ = \{s\in S \mid s\vDash^+ \f\},$ and analogously the \emph{negative extension of a formula} $ |\f|^- = \{s\in S \mid s\vDash^- \f\}$. Notice that in general a negative extension is not a set theoretical complement of the corresponding positive extension, i.e. 
$|\f|^+ \cup |\f|^-\subsetneq S$ (some states might support neither $\f$ nor $\neg\f$) and $|\f|^+ \cap |\f|^- \neq \varnothing$ 
(states in the intersection support both $\f$  and $\neg\f$). Moreover, positive and  negative extensions are mutually definable: $|\neg\f|^+ = |\f|^-$.

When we define a measure on the formulas of some language, it is natural to require that equivalent formulas have the same value (the probability of an event should not depend on its name). Another option which is technically more convenient is to work with equivalence classes of formulas directly.
Two formulas are equivalent, denoted $\phi\sim\psi$, if they are mutually derivable with respect to the axiomatisation of $\BD$ logic (see e.g.~\cite[page 5]{BilkovaFrittellaKozhMajerNazari}).
Formally, this structure is called the Lindenbaum algebra of the logic. 
For $\BD$ logic, it is the free De Morgan algebra generated by the set of atomic variables $\Prop$. It is defined as $\LTBD= (L,\land', \neg') $ such that $L$ is the set of equivalence classes $[\f] = \{\p \mid \p \sim \f\}$, $[\f] \land'[\p] = [\f \land\p]$ and $\neg' [\f] = [\neg \f]$. In what follows we will not distinguish between a formula and its equivalence class.

Information might  be incomplete or contradictory, but usually it is also uncertain. There were several attempts to propose a probabilistic version of $\BD$ logic, which extends the  basic idea of independence of positive and negative information to the probabilistic case. We build on the framework of non-standard probabilities presented in \cite{KleinMajerRad2021}. We will use the notion of a \emph{probabilistic $\BD$ model}, $\mathfrak{M}=\langle S,v^+,v^-, \mu\rangle$, which is a $\BD$ model equipped with a probability measure $\mu$ on $S$.
The probability of a statement expressed by a formula $\f$ is defined as the (classical) measure of its positive extension: $p(\f) = \mu(|\f|^+)$.
Although the non-standard probability is defined using classical measure on a boolean algebra, it satisfies axioms  weaker than the Kolomogorovian ones. In particular, the additivity axiom does not hold, it is replaced by a weaker principle called inclusion/exclusion : $p(\f \vee \p) = p(\f) + p(\p) - p(\f\land \p)$. As a consequence, some classical principles are not valid any more: it might happen that $p(\f) + p(\neg\f) < 1$ (probabilistic information is incomplete) and $p(\f\land\neg\f)>0$ (probabilistic information 
is contradictory).

\subsection{Belief and  plausibility functions}

We recall the definitions of belief functions, plausibility functions and mass functions. 
Usually those definitions are given on Boolean algebras, here, we directly generalise them to lattices.

\begin{definition}[Belief function]\label{def:generalbelieffunction}
Let $\mathcal{L}$ be a a bounded lattice. A function $\bel:\mathcal{L}\rightarrow[0,1]$ is called a~\emph{belief function} if the following conditions hold:
\begin{itemize}
\item $\mathtt{bel}(\bot)=0$ and $\mathtt{bel}(\top)=1$,
\item $\bel$ is \emph{monotone} with respect to $\mathcal{L}$: for every $x,y\in\mainL$, if $x\leq_{\mathcal{L}} y$, then $\mathtt{bel}(x)\leq\mathtt{bel}(y)$,
\item $\bel$ is \emph{weakly totally monotone}: for every $k \geq 1$ and every $a_1,\ldots,a_k\in\mathcal{L}$, it holds that 
\begin{equation}
\label{eq:pl:k:inequality}
\mathtt{bel}\left(\bigvee\limits_{1\leq i\leq k}a_i\right)\geq\sum\limits_{\scriptsize{\begin{matrix}J\subseteq\{1,\ldots,k\}\\J\neq\varnothing\end{matrix}}}(-1)^{|J|+1}\cdot\mathtt{bel}\left(\bigwedge\limits_{j\in J}a_j\right).
\end{equation}
\end{itemize}
\end{definition}

\begin{definition}[Mass function]
\label{def:general:mass:function}
Let $\mainL\neq\varnothing$ be an arbitrary lattice. A \emph{mass function} on $\mainL$ is a function $\mass:\mainL\rightarrow[0,1]$ such that $\sum\limits_{x\in \mainL}\mass(x)=1$.
\end{definition}

The following well-known lemma, see for example \cite[Theorem~2.8]{Zhou13}, shows the relation between mass functions and belief functions.
\begin{lemma}[Mass function associated to a  belief function] 
\label{lem:bel:associated:mass}
Let $\mathcal{L}$ be a finite lattice and $\bel:\mathcal{L}\rightarrow[0,1]$  a belief function. Then, there is a mass function $\mass_\bel  :  \mathcal{L} \rightarrow [0,1]$, called the \emph{mass function associated to $\bel$}, such that, for every $x\in\mathcal{L}$, 
$\bel(x)=\sum\limits_{y \leq x}\mass_\bel(y)$.
Conversely, for any mass function $\mass$ on the lattice $\mainL$, the function $\bel_\mass:\mainL\rightarrow[0,1]$ defined as $$\bel_m(x)=\sum\limits_{y \leq x}\mass(y)$$ is a belief function.  
\end{lemma}

\begin{definition}[Plausibility functions]\label{def:generalplausibilityfunction}
Let $\mathcal{L}$ be a bounded lattice. $\pl:\mathcal{L}\rightarrow[0,1]$ is called a~\emph{plausibility function} if the following conditions hold:
\begin{itemize}
\item $\mathtt{pl}(\bot)=0$ and $\mathtt{pl}(\top)=1$,
\item $\pl$ is monotone with respect to $\mathcal{L}$, 
\item for every $k \geq 1$ and every $a_1,\ldots,a_k\in\mathcal{L}$, it holds that
\begin{equation}
\label{eq:pl:k:inequality}
\mathtt{pl}\left(\bigwedge\limits_{1\leq i\leq k}a_i\right)\leq\sum\limits_{\scriptsize{\begin{matrix}J\subseteq\{1,\ldots,k\}\\J\neq\varnothing\end{matrix}}}(-1)^{|J|+1}\cdot\mathtt{pl}\left(\bigvee\limits_{j\in J}a_j\right).
\end{equation}
\end{itemize}
\end{definition}

The following two lemmas from \cite{BilkovaFrittellaKozhMajerNazari} show the mutual definability of belief, plausibility and their mass functions on De Morgan algebras.

\begin{lemma}[Plausibility function associated to a belief function]
\label{lem:bel:pl:1-bel}
Let $\mathcal{L}$ be a bounded De Morgan algebra and $\bel : \mathcal{L} \rightarrow [0,1]$  a  belief function. Then, the function $\pl_\bel : \mathcal{L} \rightarrow [0,1]$ such that $\pl_\bel(x)=1-\bel(\neg x)$ is a  plausibility function, called the  \emph{plausibility function associated to $\bel$}.
\end{lemma}

\begin{lemma}[Mass function associated to a  plausibility function]
\label{lem:pl:associated:mass}
Let $\mathcal{L}$ be a bounded De Morgan algebra, and $\pl  :  \mathcal{L} \rightarrow [0,1]$  a plausibility function. Then, the function $\bel_\pl  :  \mathcal{L} \rightarrow [0,1]$ such that $\bel_\pl(x)=1-\pl(\neg x)$ is a  belief function, called the  \emph{belief function associated to $\pl$}. We denote $\mass_\pl$ the mass function associated to $\bel_\pl$ and we call $\mass_\pl$ the \emph{mass function associated to $\pl$}. Then,
\begin{equation}
\pl(x)=1-\sum_{y\leq\neg x} \mass_{{\pl}}(y).
\end{equation}

\end{lemma}

Belief functions and their mass functions are used to reason about evidence. 
Dempster--Shafer combination rule allows merging the information provided by different sources, each source being described by a mass function.

\begin{definition}[Dempster--Shafer combination rule]Let $\mass_1$ and $\mass_2$ be two mass functions on 
$\mathcal{P}(S)$. Dempster--Shafer combination rule computes their aggregation
$\mass_1\oplus \mass_2: \P(S) \rightarrow [0,1]$ as follows.
\begin{align}
\label{eq:combination:rule:cl}
X & \mapsto 
\left\{
 \begin{aligned}
 \; &0 & \mbox{if } X=\varnothing \\
 \; &\frac{\sum \{ \mass_1(X_1) \cdot \mass_2(X_2) \mid X_1 \cap X_2 = X \} }{\sum \{ \mass_1(X_1) \cdot \mass_2(X_2) \mid X_1 \cap X_2 \neq \varnothing \} } 
 & \mbox{otherwise.}
 \end{aligned}
\right.
\notag
\end{align}
\end{definition}

\subsection{Classical updating of uncertainty measures}

A probability function $\mu$ on a Boolean algebra ${\cal P}(S)$ represents the information available about the subsets of $S$ representing events. Observing $B \in \PS$ changes the probabilities assigned to the elements of $\PS$ and leads to a new probability measure $\mu_B$. 
There are different strategies to define this new measure. Here we focus on Bayesian updating defined as $\mu_B(C)=\frac{\mu(B\cap C)}{\mu(B)}$, for every $C\in {\cal P}(S)$. In this section, we present results from \cite{Halpern2017}. We recall how Bayesian  updating is used to define conditional upper and lower probabilities, and conditional belief and plausibility, and how Dempster--Shafer combination rule can be used to define conditional belief and plausibility.
In this section, $\Pl$ denotes the plausibility function associated to $\Bel$.

\subsubsection*{Conditioning upper and lower probabilities} 
Let $\cal{A}$ be a non-empty set of probability measures defined over ${\cal{P}}(S)$. Then its \emph{lower and upper probabilities} (resp.~$\cal{A}_*$ and $\cal{A}^*$)  are functions defined on ${\cal{P}}(S)$ as follows, for every $X\in \PS $: 
\begin{equation}
    {\cal{A}}_*(X)= \inf \{\mu(X):\mu\in {\cal{A}} \} \quad\quad\text{and}\quad \quad{\cal{A}}^*(X)=\sup \{\mu(X):\mu\in {\cal{A}} \}
\end{equation}
\cite{Halpern2017} proposes the following way to update a set of probabilities $\A$ using Bayesian update. A priori, observing $B\in {\cal{P}}(S)$ leads to the Bayesian updating $\mu_B$ of all probabilities  $\mu\in\A$ such that $\mu(B)\neq 0$, that is, the probability measures consistent with that observation. 
Therefore, the update of $\A$ is defined only if there is at least one $\mu'\in \A$ such that $\mu'(B) >0$. 
We denote $\A_B=\{\mu_B:\mu \in \A \text{ and } \mu(B)>0 \}$. 
We define the \emph{conditional updating of $\A_*$ and $\A^*$ by $B$} as follows: $(\A^*)_B:=(\A_B)^*$ and $(\A_*)_B:=(\A_B)_*$.

\subsubsection*{Conditioning belief and plausibility as lower and upper probabilities}
In the classical case, \cite{Halpern2017} introduces two  ways to update belief functions: (1) via the representation of belief functions as  lower probabilities  (see Theorem \ref{the:BeliefAsLowerUpper}), and (2) via their associated mass functions (see Proposition \ref{prop:formula:DS-conditioning}).

\begin{theorem}\cite[Theorem 2.6.1]{Halpern2017}
\label{the:BeliefAsLowerUpper}
\label{BeliefAsLowerUpper}
Let $\Bel$ be a belief function defined on  $\PS$ and $\M_\Bel=\{\mu:\mu(X)\geq \Bel(X),\text{ for all }X\in \PS\}$. Then $\Bel=(\M_\Bel)_*$ and $\Pl=(\M_\Bel)^*$. 

\end{theorem}

The set $\M_\Bel$ can be updated when it contains at least one measure such that $\mu(B)>0$, that is when $\Pl_\Bel(B)>0$.
\begin{definition} Let $\Bel:\P(S)\rightarrow [0,1]$ be a belief function and $\Pl$ is associated plausibility function such that $\Pl(X)=1-\Bel(\overline{X})$ for every $X\in \P(S)$.
Then \emph{conditioning of $\Bel$ and $\Pl$  on $B$} is defined as follows:
\begin{equation*}
    \Bel_B(X)=((\M_\Bel)_*)_B(X)=((\M_\Bel)_B)_*(X) \text{ and } \Pl_B(X)=((\M_\Bel)^*)_B(X)=((\M_\Bel)_B)^*(X).
\end{equation*}
\end{definition}
 
In section \ref{sec:update plausibility}, we will work with models containing belief and plausibility functions that are not interdefinable. The following proposition characterises those functions in terms of lower and upper probabilities.
\begin{proposition}[Belief and plausibility as an upper/lower probability]\label{the:Bel:Pl:LowerUpper:2} 
Let $f$ be a function defined on $\PS$ and  $\M_f=\{\mu:\mu(X)\geq f(X),\text{ for all }X\in \PS\}$ and $\N_f=\{\mu:\mu(X)\leq f(X),\text{ for all }X\in \PS\}$. Now let $\Bel$ and $\Pl$ be  belief  and  plausibility functions defined independently on $\PS$. Then, 
\[
 \Pl=(\M_{\Bel_\Pl})^*=(\N_\Pl)^*\qquad \text{and}\qquad \Bel=(\M_{\Bel})_*=(\N_{\Pl_\Bel})_*
\]
where $\Bel_\Pl$ and $\Pl_\Bel$ are respectively the belief and plausibility associated to $\Pl$ and $\Bel$ (see Lemmas \ref{lem:bel:pl:1-bel} and \ref{lem:pl:associated:mass}). 

\end{proposition}

\begin{proof}
Recall that $\Bel_\Pl(X) = 1-\Pl(\overline{X})$.
Based on Theorem  \ref{the:BeliefAsLowerUpper}, we have 
$
\Bel_\Pl=(\M_{\Bel_\Pl})_* $ and $ \Pl=(\M_{\Bel_\Pl})^*$.
Notice that, for every $X \in \P(S)$,  
\begin{align*}
    \mu(X)\leq\Pl(X) & \iff \mu(X)\leq 1-\Bel_\Pl(\overline{X})\iff \Bel_\Pl(\overline{X})\leq 1-\mu(X)\\
    & \iff \Bel_\Pl(\overline{X})\leq \mu(\overline{X})
    \iff \Bel_\Pl(X)\leq \mu(X)\iff \Bel_\Pl(X) \leq \mu (X)
\end{align*}
Therefore, $\M_{\Bel_\Pl}=\N_\Pl$ and $(\M_{\Bel_\Pl})^*=(\N_\Pl)^*$. The proof for $\Bel$ is similar.
\end{proof}

Defining conditioning via lower and upper probabilities is not very practical from a computational perspective. The following theorem gives us an explicit formula. 
\begin{theorem}
\label{ConditioningBeliefExplicitFormula}
\label{ConditioningBeliefExplicitFormula:shortenned}
Let $\Bel : \P(S) \rightarrow [0,1]$ be a belief function. Let $\Pl$ be its associated plausibility function.
Suppose that $\Pl(B)>0$. Then, 
\begin{align*}
  \Bel_B(X) &=
 \left\{
 \begin{aligned}
 &1 & \text{ if }  \Pl(\overline{X}\cap B)=0, \\
 & \frac{\Bel(X\cap B)}{\Bel(X\cap B)+ \Pl(\overline{X}\cap B)}& \text{ if }  \Pl(\overline{X}\cap B)>0.
 \end{aligned}
 \right. \\   
 \Pl_B(X) &=
   \frac{\Pl(X\cap B)}{\Pl(X\cap B)+ \Pl(\overline{X}\cap B)}
\end{align*}
\end{theorem}
\begin{proof}
\cite[Theorem 3.8.2]{Halpern2017} proves the formula for $\Bel$ and the fact that 
\[
\Pl_B(X)=
 \left\{
 \begin{aligned}
 &0 & \text{ if } \Pl(X\cap B)=0, \\
 & \frac{\Pl(X\cap B)}{\Pl(X\cap B)+ \Pl(\overline{X}\cap B)}& \text{ if } \Pl(X\cap B)>0.
 \end{aligned}
 \right.
\] 
Notice that $\Pl(X\cap B)+ \Pl(\overline{X}\cap B) > 0$ for every $X\in\P(S)$.
Indeed, since $\Pl$ is a plausibility function, we have  $\Pl(A\cup C)\leq \Pl(A)+\Pl(C)-\Pl(A\cap C)$ for every $A,C\in \P(S)$. If $A\cap C=\emptyset$, then $\Pl(A\cup C)\leq \Pl(A)+\Pl(C)$. Hence, if $A=\overline{X}\cap B$ and $C=X\cap B$, we have $\Pl( B)\leq \Pl(\overline{X}\cap B)+\Pl(X\cap B)$. Since $\Pl(B)>0$, we have $\Pl(\overline{X}\cap B)+\Pl(X\cap B)>0$, for every $X\in\PS$. Therfore, if $\Pl(X\cap B)=0$, we have $\frac{\Pl(X\cap B)}{\Pl(X\cap B)+ \Pl(\overline{X}\cap B)}=0$ as requiered.
\end{proof}

\subsubsection*{Conditioning belief and plausibility via mass functions}
In this case, observing $B$ is encoded via the mass function $\mass_B$ as $\mass_B(B)=1$ and $0$ otherwise. The update $\Bel^B$ of a belief function  $\Bel$ by $B$ is computed via Demspter--Shafer combination rule and its associated mass function is 
$\mass_\Bel\oplus \mass_B$. To distinguish between this method and the above method we use $\Bel^B$ and $\Pl^B$ for the conditional belief and plausibility obtained by the latter approach and we call it DS-conditioning. We have the following explicit formulas for $\Bel^B$ and $\Pl^B$.

\begin{proposition}\cite[Theorem 3.8.5]{Halpern2017} \label{prop:formula:DS-conditioning}
$\Bel^B$ and $\Pl^B$ are defined if $\Pl(B)>0$. For every $X\in \PS$
\[
\Bel^B(X)=\frac{\Bel(X\cup \overline{B})-\Bel( \overline{B})}{1-\Bel( \overline{B})} \qquad \text{and} \qquad
\Pl^B(X)=\frac{\Pl(X\cap B)}{\Pl(B)}.
\]
\end{proposition}

\section{Updating of belief and plausibility over Belnap--Dunn logic}
\label{sec:update}

\subsection{Models for belief and plausibility over Belnap--Dunn logic}

We define belief functions on $\BD$ logic similarly to how we defined probabilities, that is, we use  the notion of positive/negative extension.

\begin{definition}[$\DS$ models and their associated belief functions] 
\label{def:DS:models}
Let $\LTBD$ be the Lindenbaum algebra for $\BD$ logic over the set of propositional letters $\Prop$.
A \emph{$\DS$ model} is a tuple $\mathscr{M}=
\langle S,\P(S), \Bel, v^+, v^- \rangle$ such that 
$\langle S,v^+, v^- \rangle $ is a $\BD$ model and $\Bel$ is a belief function on $\P (S)$.
We denote $\bel_\mathscr{M}^+ : \LTBD \rightarrow [0,1]$ and $\bel_\mathscr{M}^- : \LTBD^{op} \rightarrow [0,1]$ the maps such that, for every $\varphi \in \LTBD$,
\begin{align}
 \bel_\mathscr{M}^+(\varphi) = \Bel(|\varphi|^+) \qquad & \text{and } \qquad
 \bel_\mathscr{M}^-(\varphi) = \Bel(|\varphi|^-) = \Bel(|\neg\varphi|^+). 
\end{align}
We drop the subscript whenever there is no ambiguity on the model $\mathscr{M}$ we are considering.
\end{definition}
Notice that as we are defining belief of a formula via its extension, we obtain mutual definability of positive and negative belief: $\bel^-(\phi)=\bel^+(\neg \phi)$. This property mirrors how the negation works in $\BD$ logic and in non-standard probabilities.

It would be possible to define  plausibility analogously to the classical case, that is 
$\Pl(X)=1-\Bel(\overline{X})$. 
The plausibility of $\f$ would then be equal to
the sum of the masses of the sets of states that at least partially support $\f$, i.e. $\sum \{ m(A) \mid A \cap |\phi|^+ \neq \emptyset\}$. This definition does not have an intuitive interpretation, as in the sum (1) we can take into account  sets of states that all positively satisfy both $\f$ and $\neg\f$ and (2) we do not take into account sets of states that satisfy neither $\f$ nor $\neg\f$.
However, not having information about $\f$, in general, is not an argument to say that it is implausible. Therefore, we introduce models where belief and plausibility are not inter-definable.

\begin{definition}[$\DS_\pl$ models and their associated plausibility functions] 
\label{def:DS:models:pl}
Let $\LTBD$ be the Lindenbaum algebra for $\BD$ logic over the set of propositional letters $\Prop$.
A \emph{$\DS_\pl$ model} is a tuple $\mathscr{M}=
(S,\P(S), \Bel, \Pl, v^+, v^-)$ such that 
$(S,\P(S), \Bel, v^+, v^-)$ is a $\DS$ model, $\Pl$ is a plausibility function on $\P(S)$.
We denote $ \pl_\mathscr{M}^+ : \LTBD \rightarrow [0,1]$ and $ \pl_\mathscr{M}^- : \LTBD^{op} \rightarrow [0,1]$ the maps such that, for every $\varphi \in \LTBD$,
\begin{align}
 \pl_\mathscr{M}^+ (\varphi) = \Pl(|\varphi|^+) 
 \qquad & \text{and } \qquad \pl_\mathscr{M}^-(\varphi) = \Pl(|\varphi|^-)=\Pl(|\neg\varphi|^+). 
\end{align}
We drop the subscript whenever there is no ambiguity on the model $\mathscr{M}$ we are considering.
\end{definition}

In the standard approach both belief and plausibility use in fact the same information represented by the mass function, but deal with it in a different way. 
While we can see belief as the amount of information which directly supports the statement in question, plausibility represents the amount of information which does not contradict the statement. 
As Halpern says: ``$Plaus_m(U)$ can be thought of as the sum of the probabilities of the evidence that is com\-pa\-ti\-ble with the actual world being in $U$." (\cite{Halpern2017}, p. 38). 
This idea is captured in the definition of plausibility via mass function: $\pl(A) =\sum\limits_{A\cap B\neq\varnothing}\mass(B)$. 
We can also see belief and plausibility as approximations, as a lower and an upper bound for the `true' probability: $\bel(A)\leq p(A)\leq\pl(A)$. 
While in the classical case all these readings coincide, in the case of $\BD$ logic they do not, which gives us several possibilities of defining belief/plausibility pairs (see \cite{BilkovaFrittellaKozhMajerNazari} for a more detailed discussion).

\subsection{Updating belief}

A natural question that arises is what is the behaviour of the positive and negative belief functions induced by the above models, when one learns a new piece of information. Learning something about $\f$ means finding a positive or negative piece of information or even a contradictory piece of information about $\f$. Here, we directly adapt the conditioning on belief function proposed in \cite{Halpern2017}. Indeed, the belief function $\Bel$ in a $\DS$ models is defined on a powerset algebra. The non-classical behaviour with respect to the negation of $\bel^+$ and $\bel^-$ comes from the non-classical interpretation of formulas. Recall that in $\BD$ logic, $|\f|^-=|\neg\f|^+$, therefore, we only study updating with the positive interpretation of a formula.

\subsubsection*{Conditioning belief as lower  measure} If we look at belief as the lower approximation of the ``real" probability function, then we know that the ``real'' probability function is in the set $\M_\Bel$. Therefore, to update the belief after learning that $\f$ is the case, one can compute the Bayesian update of every probability in  $\M_\Bel$. In that framework, this boils down to
ignoring  information states (that is, the elements $s\in S$) not supporting $\f$. 
If $(\M_\Bel)^*(|\f|^+) = 1-\Bel(\overline{|\f|^+}) >0$, then one can define the conditional belief on $\f$ as follows: for every $X\in\P(S)$, 
\begin{equation}
    \Bel_{|\f|^+}(X)=((\M_\Bel)_{|\f|^+})_*(X)
\end{equation}
which gives us the following conditional belief function on formulas,
for each $\psi\in \LBD$, 
\begin{equation}
    \bel^+_{|\f|^+}(\psi)= \Bel_{|\f|^+}(|\psi|^+) =((\M_\Bel)_{|\f|^+})_*(|\psi|^+).
\end{equation}
In what follows, for sake of readability, we will write $\bel^+_{|\f|}$ and $\Bel_{|\f|}$ instead of $\Bel_{|\f|^+}$ and $\bel^+_{|\f|^+}$. 
Based on Theorem \ref{ConditioningBeliefExplicitFormula}, we have the following explicit formula 
\[
\bel^+_{|\f|}(\psi)=
 \left\{
 \begin{aligned}
 &1 & \text{ if }  \Bel(\overline{|\f|^+}\cup |\psi|^+)=1, \\
 & \frac{\Bel(|\psi|^+\cap |\f|^+)}{1+\Bel(|\psi|^+\cap |\f|^+)- \Bel(\overline{|\f|^+}\cup |\psi|^+)}& \text{ if }  \Bel(\overline{|\f|^+}\cup |\psi|^+)<1.
 \end{aligned}
 \right.
\]

Notice that since the update is performed on $\Bel$, both $\bel^+$ and $\bel^-$ are affected by the update.
In addition, $\bel^+_{|\f|}(\f)=1$ as expected. 
Howerver, in general, $\bel^+_{|\f|}(\neg\f)\neq 0$, because $|\f|^+\cap |\neg\f|^+ \neq \emptyset$.

\subsubsection*{Conditioning belief via mass functions}
If we interpret belief as representing the information coming from pieces of evidence, then one can also update the belief function $\Bel$ via its associated mass function $\mass_\Bel$ and Dempster--Shafer combination rule. We call that method $\DS$ conditioning.
A piece of evidence fully supporting exactly $\f$ is usually represented by the mass function $\mass_{|\f|^+}:\P(S)\rightarrow [0,1]$ such that $\mass_{|\f|^+}(|\f|^+)=1$ and  $\mass_{|\f|^+}(X)=0$ otherwise.
Therefore, the updating of $\Bel$ by finding positive information about $\f$,  denoted $(\Bel)^{|\f|}$,  is  the belief function associated to the mass function  $\mass_\Bel\oplus \mass_{|\f|^+}$. Then, based on Proposition \ref{prop:formula:DS-conditioning}, we have:
\begin{proposition} The belief function
 $(\bel^+)^{|\f|}$ is defined if $1-\Bel(\overline{|\f|^+})>0$, and, for every $\psi\in \LTBD$,
\[
(\bel^+)^{|\f|}(\psi)=(\Bel)^{|\f|}(|\psi|^+)=\frac{\Bel(|\psi|^+\cup \overline{|\f|^+})-\Bel( \overline{|\f|^+})}{1-\Bel( \overline{|\f|^+})}.
\]
\end{proposition}

It is well-known that $\DS$ combination rule is associative and commutative \cite{shafer1976}, therefore 
$\DS$ conditioning of belief functions is commutative and associative as well. 
Notice that, for every $\f,\psi\in\LTBD$, we have $\mass_{|\f|^+}\oplus \mass_{|\psi|^+}=\mass_{|\psi|^+}\oplus \mass_{|\f|^+}=\mass_{|\f|^+\cap|\psi|^+}=\mass_{|\f\wedge\psi|^+}$, which implies that $((\bel^+)^{|\f|})^{|\psi|}=((\bel^+)^{|\psi|})^{|\f|}=((\bel^+)^{|\f\wedge\psi|})$. This means that, with $\DS$ conditioning, finding both a piece of information  supporting $\f$ and a piece of information  supporting $\neg\f$ is equivalent to finding a contradictory piece of information about $\f$. Here again, notice that $(\bel^+)^{|\f|}(\neg\f)$ can be different than $0$ because some states $s\in S$ can support both $\f$ and $\neg\f$. In addition, it is worth noticing, that $\Bel_{|\f|}(X)  \leq \Bel^{|\f|}(X)$ for every $X \subseteq S$ (see \cite[Theorem 3.8.6]{Halpern2017}). Therefore, $\bel^+_{\f}(X)  \leq (\bel^+)^{|\f|}(X)$ and $\bel^-_{\f}(X)  \leq (\bel^-)^{|\f|}(X)$.

\subsection{Updating plausibility}\label{sec:update plausibility}
In some contexts it makes sense to consider the functions $\Bel$ and $\Pl$ as independent. We represent it by the notion of a $\DS_\pl$ model: $\mathscr{M}= (S,\P(S), \Bel, \Pl, v^+, v^-)$. We cannot compute the update of $\Pl$ via the update of $\Bel$ any more, however similar techniques can be used to update $\Pl$. As mentioned above, we can focus on updating based on positive information about a formula $\f$.

\subsubsection*{Conditioning plausibility as upper measure}
Based on Proposition \ref{the:Bel:Pl:LowerUpper:2}, $\Pl$ is an upper pro\-ba\-bility, that is, $\Pl=(\M_{\Bel_\Pl})^*$. So again, one can define conditioning on a formula $\f$ when $\Pl(\f)>0$ as follows: for every $\psi\in\LBD$,
 
\begin{equation}
     \pl^+_{|\f|}(\psi)=((\M_{\Bel_\Pl})_{|\f|^+})^*(|\psi|^+).
\end{equation}
and from Lemma \ref{ConditioningBeliefExplicitFormula:shortenned}, we have the following explicit formula for updating plausibilities. 

\begin{proposition}
 Let $\f$ be a formula such that $\Pl(|\f|^+)>0$,
\[
\pl^+_{|\f|}(\psi)=  \frac{\Pl(|\psi|^+\cap |\f|^+)}{\Pl(|\psi|^+\cap |\f|^+)+ \Pl(\overline{|\psi|^+}\cap |\f|^+)}  
 \]
\end{proposition}
Notice that, as expected,  $\pl^+_{|\f|}(\f)=1$ and,
since $|\psi|^-=|\neg\psi|^+$, we have:   
\[
    \pl^-_{|\f|}(\psi)=((\M_{\Bel_\Pl})_{|\f|^+})^*(|\psi|^-)= ((\M_{\Bel_\Pl})_{|\f|^+})^*(|\neg\psi|^+)=  \pl^+_{|\f|}(\neg\psi).
\]

\subsubsection*{Conditioning plausibility via mass function}
DS conditioning can be applied to plausibility functions via their associated mass functions $\mass_\pl$ (see Lemma \ref{lem:pl:associated:mass}). The mass function associated to the update of $\Pl$ based on some piece of information positively supporting $\f$ is computed via Dempster--Shafer combination rule as follows:  $\mass_\pl \oplus \mass_{|\f|^+}$.
Based on Proposition \ref{prop:formula:DS-conditioning}, we get the following formula for the corresponding plausibility function $(\pl^+)^{|\f|}$ over formulas.

\begin{proposition} 
The function $(\pl^+)^{|\f|}$ is defined if $\Pl(|\f|^+)>0$, and, for every $\psi\in\LBD$, we have
 \[
(\pl^+)^{|\f|}(\psi)=\frac{\Pl(|\psi|^+\cap |\f|^+)}{\Pl(|\f|^+)}.
\]
\end{proposition}

\section{Further work and conclusion}
\label{sec:conclusion}

This article presents methods to update belief and plausibility functions within the framework of $\BD$ logic. Recall that $\BD$ logic was introduced to reason about incomplete and contradictory information. In $\DS$ models, even though the underlying logic is non-classical, namely $\BD$ logic, the belief and plausibility functions are defined over the powerset of states. Therefore, we can import the techniques for updating belief functions from the classical logic literature. 
This work is in fact a first step. Indeed, we wish to look at belief and plausibility functions over De Morgan algebras to get a better understanding of the implication of combining belief functions and $\BD$ logic. Recall that De Morgan algebras provide the algebraic semantics for $\BD$ logic, and that there is no duality between $\BD$ models and De Morgan algebras.
Therefore, there is no way to directly import our results on frames to  De Morgan algebras.
A natural first step will be to study the mathematical properties of belief and plausibility functions over De Morgan algebras, and to establish whether they can be represented as lower and upper probabilities over sets of non-standard probabilities. This would open various options to update the belief. Indeed,  \cite{KleinMajerRad2021} presents different ways to update non-standard probabilities, among which two ways that generalise Bayesian update. 
In addition, Dempster--Shafer combination rule can straight forwardly be transferred to De Morgan algebras (see \cite{BilkovaFrittellaKozhMajerNazari}) which provides a natural way to update belief functions over De Morgan algebras. However, it remains to be checked whether this method is equivalent to $\DS$ conditioning on $\DS$ models.

\bibliographystyle{plain}
\bibliography{references.bib}
\end{document}

%% file: macros.tex
\usepackage{tikz}
\newcommand{\commentSabine}[1]{}

\usepackage{tikz}
\newcommand{\pl}{\mathtt{pl}}
\newcommand{\Pl}{\mathtt{Pl}}
\newcommand{\PS}{{\cal{P}}(S)} 
\renewcommand{\P}{{\cal{P}}}
\newcommand{\M}{{\cal{M}}} 
\newcommand{\N}{{\cal{N}}} 
\newcommand{\A}{{\cal{A}}}

\newcommand{\Prop}{\mathtt{Prop}}

\newcommand{\bel}{\mathtt{bel}}
\newcommand{\Bel}{\mathtt{Bel}}
\newcommand{\mainL}{{\cal{L}}}

\newcommand{\BD}{\mathsf{BD}}

\newcommand{\LBD}{\mathscr{L}_\mathsf{BD}}

\newcommand{\mass}{\mathtt{m}}

\newcommand{\DS}{\mathsf{DS}}
\newcommand{\LTBD}{\mathcal{L}_\mathsf{BD}}
\newcommand{\f}{\varphi}
\newcommand{\p}{\psi}